# 31th CIRP Design 2021 (CIRP Design 2021)

# Feature Visualization within an Automated Design Assessment leveraging Explainable Artificial Intelligence Methods

Raoul Schönhof*[a], Artem Werner [a], Jannes Elstner [a], Boldizsar Zopcsak [a],

Ramez Awad [a], Marco Huber [a]

[a]*Fraunhofer Institute for Manufacturing Engineering and Automation IPA, Nobelstraße 12, Stuttgart, Germany*

* Corresponding author. Tel.: +49-711-970-1843; *E-mail address:* raoul.schoenhof@ipa.fraunhofer.de

**Abstract**

Not only automation of manufacturing processes but also automation of automation procedures itself become increasingly relevant to automation research. In this context, automated capability assessment, mainly leveraged by deep learning systems driven from 3D CAD data, have been presented. Current assessment systems may be able to assess CAD data with regards to abstract features, e.g. the ability to automatically separate components from bulk goods, or the presence of gripping surfaces. Nevertheless, they suffer from the factor of black box systems, where an assessment can be learned and generated easily, but without any geometrical indicator about the reasons of the system's decision. By utilizing explainable AI (xAI) methods, we attempt to open up the black box. Explainable AI methods have been used in order to assess whether a neural network has successfully learned a given task or to analyze which features of an input might lead to an adversarial attack. These methods aim to derive additional insights into a neural network, by analyzing patterns from a given input and its impact to the network's output. Within the NeuroCAD Project, xAI methods are used to identify geometrical features which are associated with a certain abstract feature. Within this work, a sensitivity analysis (SA), the layer-wise relevance propagation (LRP), the Gradient-weighted Class Activation Mapping (Grad-CAM) method as well as the Local Interpretable Model-Agnostic Explanations (LIME) have been implemented in the NeuroCAD environment, allowing not only to assess CAD models but also to identify features which have been relevant for the network's decision. In the medium run, this might enable to identify regions of interest supporting product designers to optimize their models with regards to assembly processes.

**1. Introduction**

Around 70% of the manufacturing costs of a product are already defined during the design phase [1].

However, product developers focus their considerations on the implementation of functional features. CAD systems typically provide tools to aid in functional and performance analysis of the design [2].

Questions about further processing, on the other hand, cannot be easily extracted from the digital product image due to features that are difficult to formalize. Accordingly, optimization of the product design with regard to fitness for automation usually takes place much later in the design process or not at all. For this reason, automatically deriving abstract features such as ease of sorting, feeding and assembling, which make up considerable cost differences in automation, have so far been given not enough consideration. For example, minor component adjustments can enable the use of bowl feeders (∼ 5 t €) instead of requiring a complex bin picking robot system with image recognition for component separation in extreme cases (∼ 200 t €). Similar examples can be given for positioning and joining.

Even though abstract features can be learned from 3D CAD data by neural networks, an inherent problem remains. Relevant features are hidden within the network. This lack of feedback and interpretability prevents product optimizations.

In this paper we present a toolset comprised of the xAI Methods sensitivity analysis (SA) [3], the layer-wise relevance propagation (LRP) [4], the Gradient-weighted Class Activation Mapping (Grad-CAM) method [5] as well as the Local Interpretable Model-Agnostic Explanations (LIME) [6] for interpreting abstract three-dimensional features from CAD models previously learned in neural networks by supervised learning approaches. It is structured as followed. Chapter 2 depicts a brief overview of the previously mentioned xAI methods and their working principals. Chapter 3 then, shows how to adapt them to analyse 3D models. Chapter 4 evaluates the adapted methods briefly. The paper is completed by a short summary and future given in chapter 5 and 6.

## 2. State of the Art Explainable Artificial Intelligence Methods

Within this chapter, we aim to present a brief overview of the most relevant methods of xAI and their working principles in the domain of neural networks.

### 2.1. Sensitivity Analysis

Sensitivity Analysis aims to determine how a neural network's output is impacted by the input. It is a numeric approach to quantify the influence of every input neuron on the output [7][8]. The technique dates back to the 1960s, when the link between misclassification and weight/input perturbation of self-learning systems was first investigated [9]. Given a neural network with known weight, bias variables and architecture, one can propagate the gradient of the output layer through differentiation back through each individual neuron in the layers prior. This is calculated with respect to a loss function, which is generally set to maximize the output of a classification neuron [10]. Reaching the input layer, the gradient corresponding to the value of each input neuron can be interpreted as marking the influence of that feature. Gradient values near zero correspond to little influence on the output, whereas large ones highlight important features as given in Figure 1.

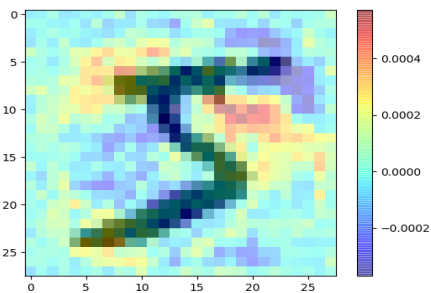

Figure 1: A CNN-Classifier trained on the MNIST dataset is used to demonstrate sensitivity analysis. Green areas: pixels (input neurons) with little influence, Red areas: positive influence, Blue areas: negative influence.

For the formal definition, the relevance or importance $R_i$ of an input $x$ is defined as [10].

$$R_i(x) = \left(\frac{\partial f}{\partial x_i}\right)^2 \tag{1}$$

The function $f(x)$ is chosen to be evidence for the true class. Decomposing of the gradient square norm yields [10]:

$$\sum_{i=1}^{d} R_i(x) = \|\nabla f(x)\|^2 \tag{2}$$

Sensitivity analysis can also be used to iteratively change an input's classification, e.g. with image classification. Once the gradients of the input layer have been determined, the additional Deep Dream algorithm [3] can be used to modify the pixel values corresponding to the gradient. The gradient is multiplied by a scaling factor and added onto the image, effectively increasing or decreasing pixel values. This output is fed back into the network to compound the modification, hence the iterative nature.

However, the limitation of sensitivity analysis is that it does not provide an explanation for the network's classification, but rather its gradient slope with respect to the input values [11]. Thus, the generated heatmap indicates pixels that make the object more or less like the target class. No interpretable information on what makes the object belong to the class is obtained.

### 2.2. Layer-Wise Relevance Propagation

During Layer-wise Relevance Propagation [12] the output of the neural network is assigned a relevancy which is propagated backward in the network.

Defining $z_{jk} = a_j w_{jk}$ as the weighted activation of neuron j onto neuron k in the subsequent layer, the relevancies of the neurons in each layer can be redistributed to each neuron j in the preceding layer following the standard LRP rule:

$$R_j^{(l)} = \sum_k \left(\frac{z_{jk}}{\sum_j z_{jk}} R_k^{(l+1)}\right) \tag{3}$$

In words, neuron j is assigned the share of the relevancy of a neuron k in the subsequent layer corresponding to neuron j's contribution to the activation of neuron k.

To enhance numerical stability, one arrives at the LRP-$\epsilon$ rule by adding a small term $\epsilon$ in the denominator.

$$R_j^{(l)} = \sum_k \left(\frac{z_{jk}}{\sum_j z_{jk} + \epsilon\, sign(\sum_j z_{jk})} R_k^{(l+1)}\right) \tag{4}$$

For layers such as convolutional layers where each input neuron is not directly connected to each output neuron via a weight $w_{jk}$, the weight has to be interpreted as the gradient $\frac{\partial a_k}{\partial a_j}$.

The propagation is implemented in four steps (see [4]).

$$z = \epsilon + layer(a) \quad Step\ 1 \quad (5)$$
$$s = R^{(l+1)}/z \quad Step\ 2 \quad (6)$$
$$c = \nabla\langle z, [s]_{cst}\rangle \quad Step\ 3 \quad (7)$$
$$R^{(l)} = a \odot c \quad Step\ 4 \quad (8)$$

Figure 2 shows an example of how to achieve this by a Tensorflow 2 implementation.

```
# a: activations, R: relevancies of all layers
# layer: function of the layer
def lrp_step(a, R, layer):
    with tf.GradientTape() as tape:
        tape.watch(a)
        z = layer(a) + epsilon
        s = R[l+1]/z
        k = tf.math.reduce_sum(z*tf.stop_gradient(s))
    c = tape.gradient(k, a)
    R[l] = a*c
```

Figure 2: Example of the relevancy redistribution between two layers of the neural network.

Applying this rule iteratively starting from the output layer until reaching the input layer, one ends up with relevancies assigned to the voxels which attempt to model their individual importance to the network classifier's decision.

There also exists an alternative framework by which to compute LRP for a wider array of network architectures, which is described in [13].

Useful out-of-the-box LRP implementations are DeepExplain [4] or iNNvestigate [14].

LRP has many desirable properties and results in heatmaps of high quality than most other methods [15].

The first among these qualities is that LRP fulfils the conversation principle, meaning that no relevancy is lost during the propagation.

$$\sum_j \left(R_j^{(l+1)}\right) = \sum_k \left(R_k^{(l+1)}\right) \quad (9)$$

Additionally, LRP heatmaps visualize features which are relevant for the network's decision, not which changes in features affect the network the most, like Sensitivity Analysis. The results obtained by LRP are also stable in the sense that small changes in the input do not significantly change the heatmaps assuming the network's decision was not altered significantly either. Lastly, LRP allow for the interpretation of both positive and negative evidence supporting/undermining the classifier's prediction.

*2.3. LIME*

The LIME algorithm proposed by Ribeiro et al. [6] is a perturbation-based method which explains predictions of a classifier by applying noise to the input data. First, a super-pixel mask is computed by using a segmentation algorithm. Then, perturbed data is computed by applying the randomly grayed out super-pixels of the mask to the input data. A super-pixel is hereby a group of pixels with the same or identical color value. The LIME algorithm passes each instance of the generated perturbations through a classifier and measures the distance of the resulting predictions to the prediction of the original input data. This distance is mapped to a value between zero and one called weight which represent similarity between the original input data and the perturbed data. Thereby, the zero represent maximal dissimilarity and the one represents maximal similarity.

Finally, a statistical method is applied on weights to explain the prediction of the classifier. Thus, there is a higher probability to approximate the amount of top features in the input data used by a classifier to make a certain decision. Because this algorithm is an optimization-based method that requires multiple iterations through a classifier, it is time and resource consuming. Ribeiro et al. describes another approach to explain the prediction of a single image on the Inception network. The first advantage of the algorithm is the classifier's independence which offers the flexibility and future-proof work with the next generation classifiers. Another advantage is the simple interpretability of the explanation. When a region of the input data is highlighted in the output, then it means that this region is important for the decision of the classifier [16].

Formally, the LIME algorithm is defined by:

$$\xi(x) = argmin\mathcal{L}(f, g, \pi_x) + \Omega(g) \quad (10)$$

where $x$ is the input data being explained, $g$ is a statistical method and $\Omega(g)$ is the complexity of this statistical method. The locality-aware loss function $\mathcal{L}(f, g, \pi_x)$ is defined by:

$$\mathcal{L}(f, g, \pi_x) = \sum \pi_x(z)(f(z) - g(z'))^2 \quad (11)$$

The proximity $\pi_x$ between the original input and perturbed data $z$ is defined by:

$$\pi_x(z) = \exp(-D(x, z)^2/\sigma^2) \quad (12)$$

where exp is the exponential kernel, $D$ is the distance metric and $\sigma$ is the width. The width defines the locality around the data.

*2.4. Grad-CAM*

The class activation mapping (CAM) method explains the prediction of a neural network by visualizing a heat map to the user. The idea behind the algorithm is that each activation map in the convolution layer preceding the GAP layer detects a different pattern in the input data. To get the heat map, the algorithm computes the sum of each recognized pattern in the activation maps. A recognized pattern, which belongs to the target class, has a greater impact on the sum. This is possible because each node in the GAP layer represents a different activation map and the weights between the GAP layer and the dense layer maps each activation map's contribution to the target class. An advantage of this method is that it requires only one iteration through the classifier [16]. This iteration consists of one forward and one backward pass which has a positive impact on computing resources and is in contrast to the perturbation-based method. One of the disadvantages of this method is the lack of interpretability [16] of the result. A description should be provided, enabling the user to get an understandable explanation. Another disadvantage is that the penultimate convolution layer should be an average pooling layer (GAP) [16].

The Grad-CAM [5] algorithm bypasses this problem by back-propagating the gradients of the last fully connected layer to the last convolutional layer, to produce a coarse localization map of important regions in the input data. The convolutional layers contain semantic and spatial information of the input data which is lost in the fully connected layers. The Grad-CAM algorithm passes the gradients of the prediction to the last convolutional producing a localization map. This localization map contains spatial information of important regions which are coarsely highlighted. The important regions for a particular layer k are defined by:

$$\alpha_k^c = \frac{1}{Z}\sum_i\sum_j\frac{\partial y^c}{\partial A_{ij}^k} \quad (13)$$

where

$$Z = \sum_i\sum_j 1 \quad (14)$$

is the number of pixels in the feature map, $\partial y^c$ is the gradient of the score for the class $c$ and $\partial A_{ij}^k$ is the gradient of feature activations $A^k$ of the convolutional layer. Finally, the localization map is defined by:

$$L^c = ReLU(\sum_k \alpha_k^c A^k) \quad (15)$$

where $ReLU$ is a rectified linear unit function. The result is a heat map with the same dimension as the convolutional layer $k$. It has to be mentioned, that Grad-CAM had been applied to 3D input and is used to assess production costs with 3D parts [17].

## 3. Proposed Work

Within this chapter, we present a range of adaptions for the previously described xAI methods as far as they are necessary. This establishes a toolbox for interpreting 3D CAD features learned by neural networks. Within the following chapters, the xAI methods have been applied to the NeuroCAD networks [18] by analysing the same sheet metal part shown in Figure 3.

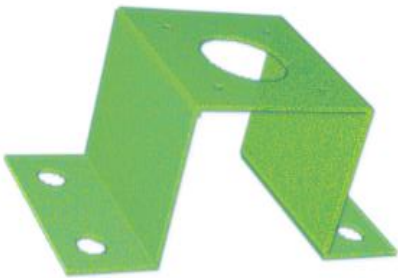

Figure 3: Sheet metal part as reference model

### 3.1. Sensitivity Analysis 3D

The sensitivity analysis can be applied for 3D voxel models of components by passing these as input through the three individual NeuroCAD networks. They assess grabability, orientability and separability of the input component. Figure 4 shows the output of the networks for a motor sheet.

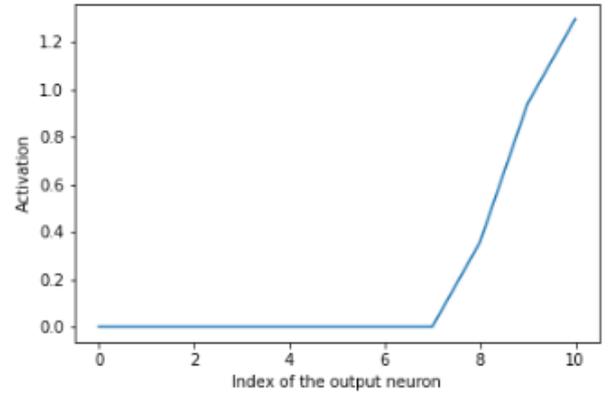

Figure 4: One can observe the plot of each of the 11 output neurons plotted against their activation. A spike in the lower, middle and upper neurons indicates a low, average and high degree of the respective assessment, in this case grabability.

To propagate the gradients back through the network, a loss function is defined to maximise the largest activation of the 11 output neurons. Figure 5 shows the heatmap of the network, showing regions of positive and negative influence to the network's classification, whereas more yellow regions are associated with a higher influence and transparent regions are associated with a lower influence.

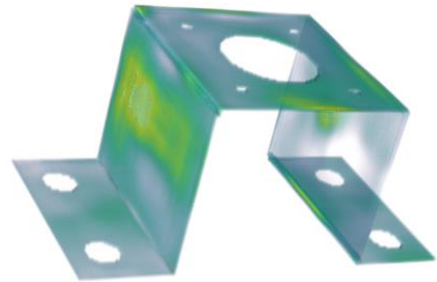

Figure 5: Heatmap for the sheet metal part, which was assessed for the abstract feature "grabability".

### 3.2. Layer-Wise Relevance Propagation 3D

When handling 3D models, Layer-wise Relevance Propagation comes very handy. It works for both 2D inputs (e.g. images) and for the 3D voxel models used for NeuroCAD without any modification to the algorithm.

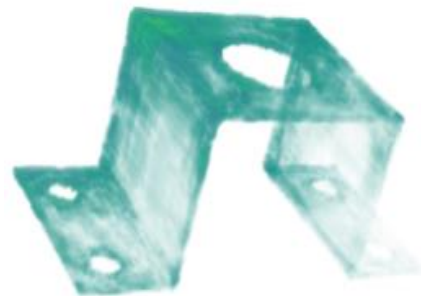

Figure 6: The corresponding LRP heatmap of the sheet metal part. The more intense the colour, the stronger the supporting evidence of the classifier's decision.

*3.3. LIME 3D*

We applied the LIME algorithm on 3D CAD data represented by point clouds to derive explanations of the classifier's predictions. In case of monotonous point clouds, we cannot use a heterogeneous segmentation algorithm, as they partition the data on different colorspace criteria. A homogeneous segmentation algorithm is also not an option, because the segmentation results in granular segments [5]. The perturbed data was created by applying the uniform three-dimensional grid to the input data as seen in Figure 6. This approach allows us to control the amount and the size of segments used to split the data which are similar to the super-pixels approach described by Ribeiro et al. and also does not require lots of computing resources. The output was predicted with a neural network and each prediction was converted to weights as described above. We use a cosine metric as a distance function to measure the proximity between the prediction of the original input data and each randomly generated perturbation. The distances were converted to weights by an exponential kernel function proposed by Ribeiro et al. The linear regression was used as the statistical method to generate important coefficients. Each coefficient represents one segment in the perturbed data. These coefficients were used to produce the amount of top features which are important for the decision of the neural network. These top features represents the classifier's prediction which is shown in Figure 7.

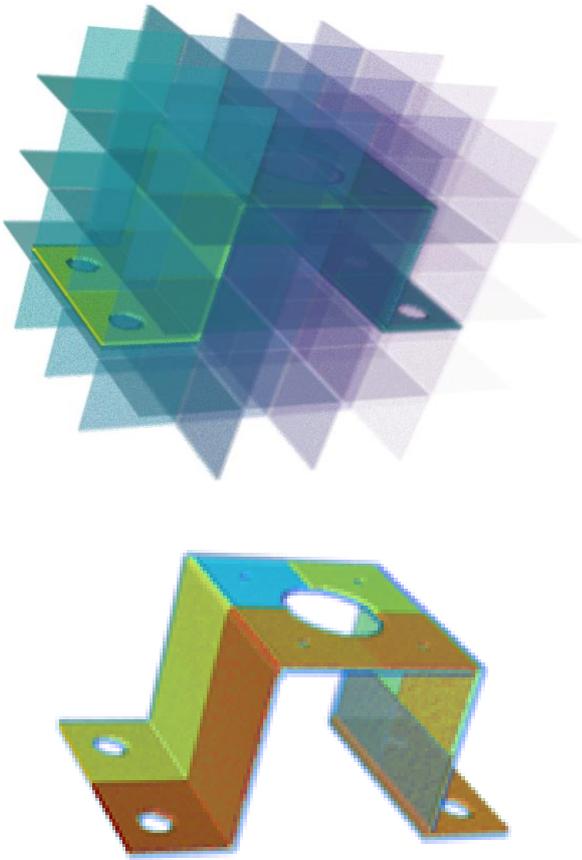

Figure 7: The uniform three-dimensional grid with 3D CAD point cloud.

*3.4. Grad-CAM 3D*

At first, we passed the input data through the classifier and grabbed the neuron activation of a given prediction. At the most, this is the top predicted class. Using Tensorflow's automatic differentiation, the gradient of the score by use of automatic differentiation with respect to the output feature map of the last convolutional layer was computed. Then, the gradients are multiplied with the activation of the feature map with respect to the predicted class. Finally, the resulting heat map is normalized and scaled to the dimension of the input data. To create a superimposed visualization, we normalize the input data and multiply it with the resulting heat map. The output is the voxel model representing a 3D CAD data which visualize regions that are important for the prediction with higher byte values and regions that are irrelevant with lower byte values. This output is shown in Figure 8 by applying a colour map in order to improve the interpretability for a user.

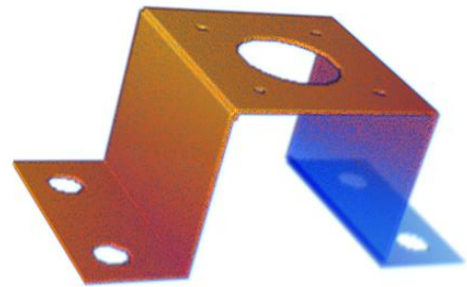

Figure 8: The result of Grad-CAM algorithm. The yellow or orange colour space represents the relevant regions. The blue colour space represents irrelevant regions.

**4. Evaluation**

For evaluation, a simple 3D CNN was created. This was done by first pretraining an autoencoder unsupervised. Then, a supervised transfer learning with labelled data was performed.

*4.1. Autoencoder Pre-Training*

The current NeuroCAD trainset for abstract manufacturing features is quite small (approx. 200 models) [18], which is why only relying on model augmentation can lead to an overfitting. To overcome this issue, a large part of the network was pretrained by using an autoencoder. This allowed to use unsupervised learning approaches and large, already existing, trainsets. For pretraining, the autoencoder was trained by 10.000 models from the ABC trainset [20].

*4.2. Transfer learning on Abstract Features*

Afterwards, a transfer learning based on the NeuroCAD trainset was applied allowing the network do distinguish between three abstract features from assembly automation: separability, grabability and orientability. These values have been assessed by expert interviews, representing the associated fitness from an automation expert's perspective [18]. Its decisions have been visualized and interpreted by the adapted 3D xAI Methods presented in this work.

*4.3. Results*

An overview of the different xAI method outputs is depicted in Table 1.

Table 1. Summary: Outputs of the 3D xAI methods, LIME 3D, Grad-CAM 3D, SA 3D and LRP 3D. The networks analysed had been taken from the NeuroCAD project [18].

| Method | Grabability | Seperabitily | Orientability |
|---|---|---|---|
| LIME/Jet | 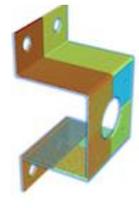 | 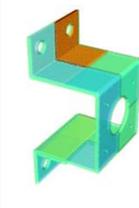 | 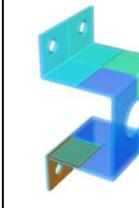 |
| Grad-CAM | 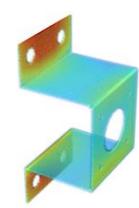 | 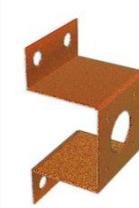 | 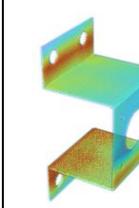 |
| Sensitivity Analysis | 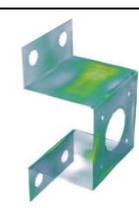 | 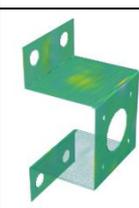 | 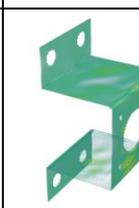 |
| LRP | 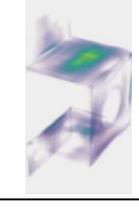 | 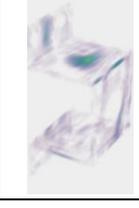 | 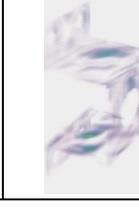 |

## 5. Conclusion

Within this work, a toolbox of the most current xAI Methods have been presented, which have been adapted to support 3D inputs. Thus enabling the users to analyse neural network based decisions made on 3D voxel models. Comparable to 2D image analysis, these tools will allow the interpretation of complex-abstract features assessed by neural networks.

## 6. Future Work

As already stated, the 3D xAI methods, only allow to highlight regions of the CAD part which are especially relevant to neural networks. It does not give the user the interpretation. Therefore the next task, will be to create a metrology how to interpret the generated results. Being able to assess neural network based decisions and identify regions in 3D objects could be of interest especially for product designers in order to improve their designs. Also the presented toolbox was tailored for handling voxel models. Anyway, there is no reason not to apply the methods on any other 3D model representation. Especially the investigation of point cloud models could lead to the direct assessment of scanned models, leveraging 3D assessments of physical parts.

## 7. Acknowledgements

The research presented in this paper has received partial funding by the AI-Innovation Centre "Learning Systems" (KI-FZ) (10/2019 - 03/2021).